\documentclass[dvipsnames,format=sigconf]{acmart}

\usepackage{tabularray}
\UseTblrLibrary{siunitx}

\usepackage{todonotes}

\AtBeginDocument{%
  \providecommand\BibTeX{{%
    \normalfont B\kern-0.5em{\scshape i\kern-0.25em b}\kern-0.8em\TeX}}}

\copyrightyear{2023}
\acmYear{2023}
\setcopyright{rightsretained}
\acmConference[GECCO '24]{Genetic and Evolutionary Computation Conference}{July 14--18, 2024}{Melbourne, Australia}
\acmBooktitle{Genetic and Evolutionary Computation Conference (GECCO '24), July 14--18, 2024, Melbourne, Australia}

\begin{document}

\title{Stitching for Neuroevolution: Recombining Deep Neural Networks without Breaking Them}

\author{Arthur Guijt}
\email{Arthur.Guijt@cwi.nl}
\orcid{0000-0002-0480-2129}
\affiliation{%
  \institution{Centrum Wiskunde \& Informatica}
  \city{Amsterdam}
  \country{The Netherlands}
}

\author{Dirk Thierens}
\email{D.Thierens@uu.nl}
\orcid{0000-0002-9308-5159}
\affiliation{%
  \institution{Utrecht University}
  \city{Utrecht}
  \country{The Netherlands}
}

\author{Tanja Alderliesten}
\email{T.Alderliesten@lumc.nl}
\orcid{0000-0003-4261-7511}
\affiliation{%
  \institution{Leiden University Medical Center}
  \city{Leiden}
  \country{The Netherlands}
}

\author{Peter A.N. Bosman}
\email{Peter.Bosman@cwi.nl}
\orcid{0000-0002-4186-6666}
\affiliation{%
  \institution{Centrum Wiskunde \& Informatica}
  \city{Amsterdam}
  \country{The Netherlands}
}
\affiliation{%
  \institution{Delft University of Technology}
  \city{Delft}
  \country{The Netherlands}
}



\begin{abstract}
  Traditional approaches to neuroevolution often start from scratch. This becomes prohibitively expensive in terms of computational and data requirements when targeting modern, deep neural networks.
  Using a warm start could be highly advantageous, e.g., using previously trained networks, potentially from different sources. This moreover enables leveraging the benefits of transfer learning (in particular vastly reduced training effort).
  However, recombining trained networks is non-trivial because architectures and feature representations typically differ.
  Consequently, a straightforward exchange of layers tends to lead to a performance breakdown.
  We overcome this by matching the layers of parent networks based on their connectivity, identifying potential crossover points. To correct for differing feature representations between these layers we employ stitching, which merges the networks by introducing new layers at crossover points. To train the merged network, only stitching layers need to be considered. New networks can then be created by selecting a subnetwork by choosing which stitching layers to (not) use. Assessing their performance is efficient as only their evaluation on data is required. 
  We experimentally show that our approach enables finding networks that represent novel trade-offs between performance and computational cost, with some even dominating the original networks.
\end{abstract}

\definecolor{stitchcolor}{HTML}{56517e}
\definecolor{switchcolor}{HTML}{6c8ebf}
\definecolor{additionalcolor}{HTML}{b7a571}
\definecolor{choicecolor}{HTML}{82b366}



\maketitle

\section{Introduction}\label{sec:introduction}
In recent years, deep neural networks have shown to be incredibly powerful machine learning models in many fields. These neural networks consist of parameterized layers performing operations on their input(s), and providing an output. A subset of these parameters are the architectural parameters, which for example include the depth of the network, the number of input and output features of a linear layer, the and number of channels or the kernel size of a convolution layer.

Rather than setting these parameters by hand, Neural Architecture Search (NAS) aims to automate this process by searching for the best settings of these architectural parameters. As these parameters potentially change the topology of the network and the number of weights to be trained, the easiest way to accomplish this is by training a deep neural network from scratch each time a new parameter setting is considered. However, doing so takes a significant amount of time and resources for even a single choice of parameters.

Supernetworks avoid re-training the network at every evaluation by creating a very large compound network, which contains each possible architectural choice, such that each layer is already trained~\cite{caiOnceAllTrain2020}. When evaluating a specific architecture using a supernetwork, there is no longer a need to train the network pertaining to this specific architecture. Instead, only the performance of a subnetwork in the supernetwork that represents this architecture needs to be assessed.

In general, the cost of training a deep neural network is often reduced by using pre-trained networks. By taking a network trained on a different task or dataset and employing transfer learning~\cite{caruanaMultitaskLearning1997,panSurveyTransferLearning2010,neyshaburWhatBeingTransferred2020}, a specialized network is constructed which can re-use knowledge gained from training on this task or dataset. By doing so, both the training costs and the necessary amount of data can be reduced drastically, while simultaneously also increasing the performance of the resulting network for the new task.
In~\cite{munozAutomatedSuperNetworkGeneration2022} such pre-trained networks were converted into supernetworks using specifically crafted rules in order to vary specific architectural parameters. However, the networks resulting from such a procedure are limited to simplified, pruned versions of a single network.  
In a (neuro-)evolutionary setting, one would like to be able to recombine parts of multiple networks, reusing and combining parts that have been known to work well.
Given that supernetworks have been used to efficiently search for new network architectures, a similar approach could be utilized to obtain an efficient recombination operator. In other words, can we create a supernetwork that allows one to vary between combinations of two pre-trained neural networks, while preserving or even improving upon these pre-trained neural networks?

Recombining neural networks is a difficult problem due to the compatibility issues that appear between different networks, as will be discussed later. These issues are only compounded by the presence of multiple, different architectures. In this work, we study a new approach to do recombination that overcomes these issues. Below, in Section~\ref{sec:recombine}, we first describe relevant background and related literature.

\section{Recombining Neural Networks}\label{sec:recombine}
Recombining two neural networks by, e.g., crossover, without breaking them is a nontrivial task~\cite{stanleyEvolvingNeuralNetworks2002, uriotSafeCrossoverNeural2020, ainsworthGitReBasinMerging2023}.
This is because of representational differences, which we will separate into two categories. First, there is the problem that different parts of the network may encode different features. For example, for vision tasks features in the first layers of a network are generally lower level, representing aspects such as edges, whereas features represented in later layers include higher level features - such as eyes or other complex objects~\cite{olah2017feature}. Providing the wrong features to a layer, will cause the resulting features to be considerably different, and in all likelihood to become unusable. To perform crossover without this disruption, one will need to know \emph{where} matching features are.
Second, even if similar features are present in a part of the network, they may be represented differently. This is caused by structural-functional redundancies induced by the networks' structure, also referred to as the competing conventions problem~\cite{schafferCombinationsGeneticAlgorithms1992,thierensNonredundantGeneticCoding1996,stanleyEvolvingNeuralNetworks2002}. This problem concerns, for example, the ordering of features in a hidden layer of a Multi-Layer Perceptron (MLP) or classical neural network. Here, the weights of the corresponding and next layer can be permuted to reorder the features without affecting the end result. This causes a permutation symmetry among the weights.
During initialization and training symmetries, such as this ordering property, will be determined at random, causing different networks to almost certainly adhere to different feature representations.

Contrary to MLPs, which consist of sequences of linear layers and their activation functions, modern deep neural networks consist of a large variety of layers, potentially with parallel branches. It is no longer a given that the permutation symmetry holds for each kind of layer used in practice. By extension, different layers may also introduce new kinds of symmetries. Generally, given two neural networks with different architectures, the correspondence between the two networks becomes unclear. We can no longer operate under the assumption that the same layer fulfills the same role, for there is no longer a notion of positional equivalence. 

Yet, for neuroevolution we would want to have the means to recombine a large variety of deep neural networks, including different architectures with parallel branches. Performing crossover between two such deep neural networks therefore requires both (1) knowledge about matching layers and (2) the means to align their representation. In literature, (1) is either bypassed by assuming identical architectures and assuming positional equivalence, or by tracking transformations as a network is grown~\cite{stanleyEvolvingNeuralNetworks2002}.
Such approaches are not helpful here: assuming positional equivalence only works if the architectures are the same, and hence does not work across different architectures; and tracking as the network grows is a useful technique for newly introduced layers, but does not allow for connections to be found between pre-existing networks.
Common strategies to tackle (2) are to permute the neurons, so that their activations match up~\cite{uriotSafeCrossoverNeural2020, ainsworthGitReBasinMerging2023}, or by selecting and matching individual features and merging layers~\cite{stoicaZipItMergingModels2023}. Yet, these approaches are restricted to correcting for the ordering, inclusion, and absence of features, and may miss some layer-specific symmetries. An example of such a symmetry is the use of the sigmoid activation function~\cite{thierensNonredundantGeneticCoding1996}, where due to the symmetry of the activation function, the sign of the feature may be flipped.

In this work, we will investigate the use of model stitching~\cite{bansalRevisitingModelStitching2021} for network recombination. Model Stitching was originally intended to assess feature map similarity, much like canonical correlation analysis~\cite{raghuSVCCASingularVector2017,morcosInsightsRepresentationalSimilarity2018} and centered kernel analysis~\cite{kornblithSimilarityNeuralNetwork2019,williamsGeneralizedShapeMetrics2021}. Model stitching essentially connects the output of one layer to the input of another layer in another network. Yet, model stitching has one major difference with naively making these connections directly: a simple trainable layer is inserted in between the layers to be connected. The parameters of this intermediate \emph{stitching layer} are then determined by training the network with all other parameters frozen.
This newly introduced stitching layer serves as a translator and corrects for the various misalignments and symmetries present in the representation of the layers to be connected.
The chosen kind of stitching layer is important: a complex layer that can perform feature extraction, while powerful, yields more expensive networks and is costly to train. Yet, a layer that is too simple cannot correct for all desired symmetries, e.g., an affine transformation cannot reorder features.
As such, in this work, we will focus on linear, or linear-like layers, such as convolutional layers with a 1x1 kernel, without an activation function.


\section{Efficient Model Stitching}\label{sec:efficientmodelstitching}
\begin{figure*}
  \centering
  \includegraphics[width=0.9\textwidth]{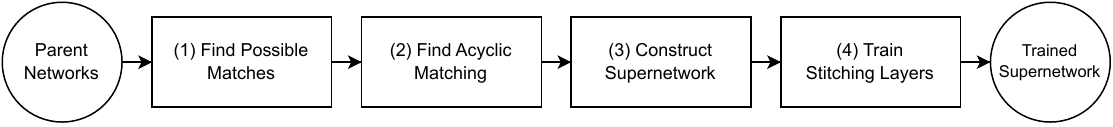}
  \caption{An overview of the process to generate a single supernetwork based on two parent networks.}
  \label{fig:overview-stitching}
\end{figure*}

\begin{figure}
    \centering
    \includegraphics[width=0.9\columnwidth]{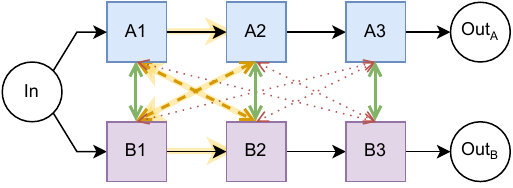}
    \caption{
      Given two networks A and B, we consider matching their layers. Some transformations may be impossible given the kinds of stitching layers provided.
      Furthermore, some matches may preclude others as they would result in the graph containing cycles after introducing the stitch and switch layers, for example the edges dashed in orange preclude one another. Green edges represent a valid matching.
    }
    \label{fig:matching}
\end{figure}

The following paragraphs explain how we construct a supernetwork that preserves the layers and weights of two parent networks, while allowing for the exchange of layers between the two networks. For a schematic representation of the steps, refer to Figure~\ref{fig:overview-stitching}. The procedure starts with a pair of parent networks. The networks used within this work and how they are imported is listed in Section~\ref{sec:networks}.

\paragraph{Finding possible matches}
Before we can start combining the networks using stitching layers, note that the chosen kinds of stitching layer can restrict which stitches are possible; there needs to be a layer able to take both the right kind of input and produce the right kind of output. In this work, we use two kinds of stitching layers, which translate between the output of given a pair of layers $A$ and $B$ (and vice-versa). Both kinds require that the output of both of these layers are tensors of some shape $sh_A$ and $sh_B$, which provide the size of each dimension.

First, we consider stitching using a linear layer if the dimensionality of the tensor $\text{dim}(A) = \text{dim}(B) = 2$. We label the dimensions of the tensor, $sh_A = [B, n]$, $sh_B = [B, m]$, where $n$ and $m$ represent the number of features per sample produced by this layer, and $B$ the batch dimension. The layer is then configured to be a linear layer taking $n$ features and producing $m$ new features.
Second, a stitching with a Conv2D layer is considered if $\text{dim}(A) = \text{dim}(B) = 4$, $sh_A = [B, n, w, h]$, and $sh_B = [B, m, w, h]$, where $w$ and $h$ represent the \emph{identical width and height} of the last two dimensions of both tensors. Here we create a stitching layer that takes $n$ channels and produces an output with $m$ channels. 
In the remainder of this process we can only match two layers $A$ and $B$ if one of the layers above can convert between the output of layers $A$ and $B$. However, this is not the only restriction.

\paragraph{Finding an acyclic matching}
Considering all possible stitches for transformation simultaneously turns out to almost always be impossible: this would result in a cycle in the constructed network. Such a network is unusable, as there exists no ordering to perform a feed-forward pass.
The orange dashed edges in Figure~\ref{fig:matching} show an example of a matching that would cause a cycle, which has been highlighted.
In this example, after the transformation, the input of A2 would either originate from A1, or from B2 (using the stitch). Similarly, B2 can obtain its value from B1 or A2. A2 can therefore be dependent on itself for input. 
Without an ordering for a feed-forward pass, stitching layers cannot be trained simultaneously, nor are subnetworks guaranteed to be valid. Therefore, we must limit ourselves to a matching that does not generate cycles in the resulting graph. To do so, we solve a matching problem with ordering constraints using a branch and bound approach as described in the supplementary material, using a metric that maximizes the number of matches between two networks. The resulting acyclic matching under these constraints indicates where similar features are based on the structure of the network.


\paragraph{Constructing supernetwork}
Given a set of matchings, pairs of layers $A$ and $B$ (and vice-versa) for which a kind of stitching layer is defined, we can merge and transform the networks into a supernetwork. This transformation introduces stitching layers to transform the output of layer $B$ into what the output of layer $A$ is expected to be, and a switch $s_A$ which allows for the selection of the output. The original input and the result of the stitch are provided as input to the switch, such that the switch can select either the original output of $A$ or the output of the aforementioned stitch. All original connections to the output of $A$ are replaced with the output of this switch $s_A$. By convention, we ensure that the first input to any switch represents the original output of the corresponding layer.
Finally, the output is a special case. For the output no stitch is necessary as the format of the output is defined by the task itself. Given identical tasks the output is directly connected to the output switch, without any additional stitching layer. For example, using the (bidirectional) pairs given by the green bidirectional arrows in Figure~\ref{fig:matching}, one would obtain a network similar to the one illustrated in Figure~\ref{fig:stitchnet-diagram}. 

In addition to picking one of the outputs of the two networks, we include the option of creating an ensemble using these outputs as a third input to the output switch.
While ensembles are often better performing than their constituent networks, the computational cost is that of both networks added together. Yet, the newly introduced stitches allow for offspring to share a subset of the layers, reducing the computational complexity of the resulting ensemble accordingly.


\paragraph{Training stitching layers} Finally, given a network with no cycles, we perform a simultaneous training procedure for each stitch. This procedure is applied \emph{once} as part of network creation and trains the weights of \emph{all stitching layers}. The resulting weights are used by all evaluations such that no training need to take place when evaluating a subnetwork.

First, all layers except for the stitching layers are frozen.
We then train by adding together the Mean-Squared Error (MSE) loss at each switch, which is calculated between the output provided by the stitch, and the output of the original layer as target. As all layers of the original networks are frozen, only the stitches require a gradient. The output of the stitch is only used to compute the MSE loss immediately, and does not affect the result of the network, including any other stitching layers. This allows us to train all stitching layers using the same forward and backward pass with little additional computational cost.
This especially reduces computational costs compared to the original approach described in~\cite{bansalRevisitingModelStitching2021}, where each stitching layer has its own \emph{individual} task-specific (supervised) training procedure. 
In addition, while in the original approach the training of a stitch was prone to failure when there exist significant differences in the mean and variance of the activations of the layers to be stitched, the newly proposed approach can successfully train such stitching layers, due to the gradient being much better behaved in this case.

\section{Stitching for Neural Architecture Search}\label{sec:stitching-nas}
\begin{figure*}[hbt]
    \centering
    \includegraphics{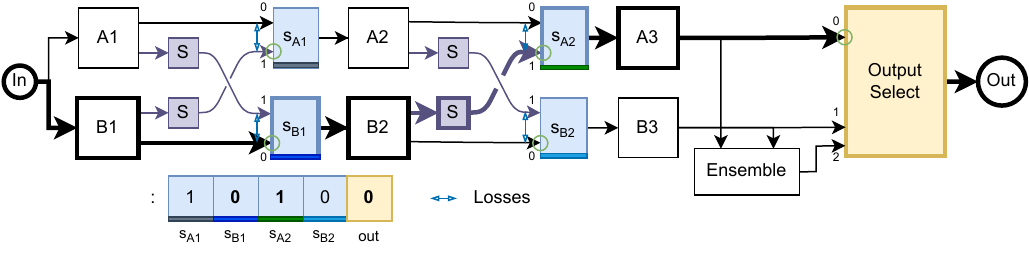}
    \caption{A diagram representing the supernetwork generated for the example in Figure~\ref{fig:matching}, consisting of parent layers, \textcolor{stitchcolor}{stitches}, and \textcolor{switchcolor}{switches}. The fixed length genotype is a discrete decision vector which for each switch determines which input to take for each switch, and have for the given example been marked with \textcolor{choicecolor}{green circles}. By working backwards from the output, taking only the selected input in the case of a switch, a subset of the layers and paths is selected. The corresponding layers and arrow for the example representation has been made bold. The \textcolor{additionalcolor}{output layer} is a switch which selects between the networks, or the ensemble thereof.  }
    \label{fig:stitchnet-diagram}
\end{figure*}

From the process in Section~\ref{sec:efficientmodelstitching} we obtain a supernetwork consisting of the layers of both parent networks, as well as the newly introduced stitch and switch layers that create new connections between the parent networks. An example of such a network is visible in Figure~\ref{fig:stitchnet-diagram}. A switch picks the source of its output in the supernetwork. By deciding for every switch which input to take ('original' or 'stitched') we can select a subnetwork. 
We combine the decision for each switch into a fixed-size discrete representation.

Using these decisions, the subnetwork can be obtained by working backwards from the output, simplifying each switch to the selected input connection. Layers that are not encountered, and therefore no longer affect the output, are then removed. This results in a subnetwork that can be evaluated, without the requirement of additional training. For instance, by assessing the performance of a neural network on a validation set and/or by determining the number of computational operations required (e.g., multiply-adds).


While the sequential nature of a neural network generally requires us to re-evaluate a network from scratch every time, not all changes to the decision variables affect the chosen subnetwork. This is because some switches are inactive: their output is unused. For such a switch changing which input is used does not result in any change in the subnetwork that has been selected. By tracking which switches were active during a previous evaluation, and allowing this information to transfer to the offspring during variation, we can detect whether all changes are restricted to variables belonging to inactive switches.
When all changes have occurred to inactive variables, the network has not changed - and the original objective value is still valid, allowing us to skip the evaluation step.




\section{Studying the space of offspring}\label{sec:investigating-search-space}
By stitching the two networks together, we are able to select combinations of the two networks according to a representation. While the idea is to use this as a crossover operator within a larger neuroevolutionary scheme, for this work restrict ourselves to the investigation of the search space of potential offspring. In this work, we will be using this crossover representation as a genotype, allowing us to search through the space of potential offspring. This provides us with knowledge about what kind of offspring can be found, and allows us to obtain results on what kind of search approach works best for obtaining good offspring according to some criteria.

The search will be multi-objective, as we wish to find networks that are both more performant, i.e., have higher accuracy, while minimizing the corresponding computational cost, i.e., number of multiply-adds. This computational cost is important, as creating a standard ensemble may perform well, it is also costly to use. Having the computational cost as a second objective provides pressure to merge redundant computation between the stitched networks where possible.


Despite the training procedure, a stitching layer may be unable to fully perform the desired transformation, introducing some error. Such a stitching layer may end up having a negative impact on the performance of any network using it. The error introduced by such a stitch may cause further layers and stitches to additionally increase the error further. If stitches are sampled uniformly, there exists a high likelihood that the performance is reduced by such stitches, potentially to the point that the network becomes nonfunctional.

To avoid starting with a pool of non-functional initial networks due to this issue, we initialize the genotype such that the resulting networks use fewer stitches. The networks without any stitches are the original networks, the genotypes corresponding to the original networks consist of all zeroes by convention, excluding the output. This knowledge is used by performing initialization (or sampling) in a biased fashion. Let $1 - p$ be the probability to preserve original network links (and sample a 0), and $p$ the probability of using the stitch at each switch. We performed a preliminary experiment using the first network pair, ImageNet (a), testing the values $p \in \{0.02, 0.10, 0.50\}$ to cover varying scales, we determined that $p = 0.02$ worked better than $0.10$ or $0.50$.
As we hypothesize that the number of stitches used has a strong impact on the performance of a network, independent of string length, we pick $p$ such that the number of 1's in the genotype is expected to be the same. i.e., as for ImageNet (a) the string length is $\ell=309$, the number of expected 1's is $0.02 \times 309 = 6.18$. To keep this number identical across network pairs, we use $p = 6.18 / \ell$, where $\ell$ is the string length.

While the cost of evaluating a network has been reduced greatly, it is still in the order of seconds because a network must still be applied to each element in the dataset.
To speed up the search, we use asynchronous parallel evolutionary algorithms, such that we can maximally utilize the available computational resources.
All the Evolutionary Algorithms (EAs) are made asynchronous parallel by having a loop for each of the $n$ individuals in the population: generate new solution, request evaluation (add to queue), wait for evaluation to complete, process solution (e.g., selection, update archive), repeat. Consequently, there are approximately $n$ solutions pending evaluation at any given time.

Finally, in preliminary experiments, we found that for ImageNet (a) there are low performance, yet low computational cost networks on the approximation front, and have accuracies ranging from $0.10$ to $0.30$. In practice, the most interesting networks are often found at the higher end of the performance scale. Finding these low performance networks consumes a significant portion of the computational budget. To ensure that EAs allocate a larger portion of the budget towards finding high-performing networks, we utilize a constraint annealing scheme similar to Adaptive Steering~\cite{alderliestenGettingMostOut2015} to steer the search towards higher accuracy values. This is implemented by introducing a threshold with respect to accuracy. This threshold linearly increases to $0.5$, to gradually exclude these low performance solutions with some margin, until half the available budget has been used. Solutions below this threshold are considered always strictly worse than solutions above this threshold, even if other objectives would otherwise have allowed this solution to be a part of the approximation front.

Source code and data will be made available.

\subsection{(LK-)GOMEA}
The interlinking of layers via stitches and switches impose an intricate structure. Depending on which stitching layers are the most useful, we expect patterns to emerge, indicating stitches which are used jointly. Linkage learning, which infers structure from the population, may be a suitable technique in order to infer and exploit this structure to improve performance. We therefore utilize one such linkage-learning EA, Gene-pool Optimal Mixing Evolutionary Algorithm (GOMEA)~\cite{dushatskiyParameterlessGenepoolOptimal2023}. In short, GOMEA employs a linkage tree, which is learned using UPGMA based on a mutual information matrix. This linkage tree contains sets of genotype indices, which are deemed linked. These indices are employed in a local-search like operation named Gene-pool Optimal Mixing (GOM) where for each of these indices, values are sampled from the population, replaced, and tested for improvement. We implement and apply an asynchronous parallel variant similar to that employed in~\cite{guijtImpactAsynchronyParallel2023} with improvements judged similar to MO-GOMEA using only Tschebysheff scalarizations~\cite{luongImprovingPerformanceMORVGOMEA2018}. Scalarization weights are reassigned by collecting and, in random order, assigning each of the weights to the individual with the lowest scalarization without newly assigned weights. This reassignment occurs after every $n$ applications of GOM.

For the given problem, set of active variables changes depending on the values assigned to the variables themselves. The relationships and dependencies between variables can therefore be different in different parts of the search space. This presence of multiple linkage structures has shown to be detrimental to the performance of linkage learning approaches~\cite{guijtSolvingMultistructuredProblems2022}. Therefore, we also investigate a linkage-kernel variant of GOMEA (LK-GOMEA) that was proposed to be more robust to such variable linkage structure.
For LK-GOMEA the linkage is learned individually for each solution over its k-nearest neighbors. Due to the absence of a generational clock, we cannot use the original population sizing scheme. As $k$ was tied to this scheme, we determine $k$ differently. A value for $k$ is chosen by randomly sampling a number of modes for each solution $m_i$ uniformly between $1$ and $n / c = m_\text{max}$, where $c$ is the minimum neighborhood size. Then, $k = \left\lceil n / m \right\rceil$. This per kernel neighborhood size should be more likely to sample a small value, as the expected value is $\log_2(n)$, yet also have the possibility to sample a larger neighborhood to avoid premature convergence.

\subsection{Genetic Algorithm}
Due to the complicated structure of the problem, the linkage learning approach may fail to exploit the dependencies successfully. We therefore include an asynchronous genetic algorithm (GA) which utilizes a general notion of structure present in the genotype.
We create new offspring by performing two-point crossover and uniform mutation with $p=\frac{1}{\ell}$, where $\ell$ is the string length. 
Two-point crossover was chosen because the variables are in a topological order of the neural network. Furthermore, in preliminary experiments it was found to perform at least as well as one-point crossover.
Due to the asynchronous nature of the EA we use a replacement-like scheme, similar to the one used in MOEA/D~\cite{zhangMOEAMultiobjectiveEvolutionary2007}, but restricted to the parents to preserve diversity. As such, selection is performed by randomly selecting a parent whose scalarized fitness is worse, and replacing it. The offspring is not selected if no such solution exists. Scalarization is performed similarly to GOMEA, with the exception that scalarization weights are reassigned after every $n$ completed evaluations.


\section{Networks}\label{sec:networks}
We perform experiments with respect to two tasks - classification on ImageNet/ImageNetV2~\cite{russakovskyImageNetLargeScale2015,rechtImageNetClassifiersGeneralize2019} using pre-trained networks from timm~\cite{rw2019timm} and Semantic Segmentation on the PASCAL VOC Dataset~\cite{Everingham10} using networks from TorchVision~\cite{TorchVision_maintainers_and_contributors_TorchVision_PyTorch_s_Computer_2016}. These models are provided as code, rather than a graph representation. During a sample execution we keep track of which module provides input to what other module. Providing us the graph corresponding to the network.
To ensure correctness, for each network imported, we have verified the output of the converted network to be the same as that of the original network on some sample input.
Using these imported networks we create stitched networks following the procedure described in Section~\ref{sec:efficientmodelstitching}. For each dataset used, the train-validation-test split, and the input networks used for stitching differ, and are described in the following paragraphs.

\paragraph{ImageNet} The data-splits used for the ImageNet classification task are as follows. The training set is the training set of ImageNet~\cite{russakovskyImageNetLargeScale2015}, using only the samples that also have bounding box annotations ($546\,545$ out of $1\,283\,166$ images in the full dataset) validation is done using all unique samples of ImageNetV2~\cite{rechtImageNetClassifiersGeneralize2019}, the test set used here is the validation set of ImageNet~\cite{russakovskyImageNetLargeScale2015}. The stitching layers are trained using Adam~\cite{kingmaAdamMethodStochastic2017} with a learning rate of $lr = 10^{-3}$ and a batch size of $32$ chosen to maximize hardware utilization, using $16\,384$ random samples (out of $546\,545$ samples), from the training set - noting that training has generally converged at this point. For evaluation, only the first $1000$ samples of the validation set (out of $50\,000$) are used to keep evaluation costs lower, without going below the number of classes for this problem.

On ImageNet we consider stitching the following two pairs of networks, all originating from the timm model library~\cite{rw2019timm}. The first pair (a) consists of resnet152~\cite{heDeepResidualLearning2016} and efficientnet\_b4~\cite{tanEfficientNetRethinkingModel2019}. The resulting network has a total of $154$ matches, or $\ell=2 \times 154 + 1 = 309$ switches and corresponding decision variables. With the available computational resources for this experiment, we were able to evaluate up to $90$ solutions in parallel at a given time, evaluating up to $5$ solutions simultaneously per GPU with a batch size of $32$. Creating this supernetwork took a total of $1415.52s$. The majority of which is due to training the stitching layers ($994.28s$, step 4 in Figure~\ref{fig:overview-stitching}). Other steps (1-3) added up to a total of $421.24s$, most of which is due to finding an acyclic matching ($417.02s$, step 3).

The second pair (b) consists of resnet50~\cite{heDeepResidualLearning2016} and resnext50\_32x4d \cite{xieAggregatedResidualTransformations2017}, which are two architectures that are more similar, with the latter modifying the repeated block by grouping channels. Stitches are trained identically to (a). The resulting network has $206$ matches, resulting in a total of $413$ decision variables. Stitching these networks took $1463.62s$, most of which was training the stitching layers ($1460.24s$, step 4). With the available computational resources for this experiment, we were able to evaluate up to $72$ solutions in parallel at a given time, or $4$ / GPU.

\paragraph{VOC} To show that this stitching approach can be applied to other tasks with different architectures, we include a different kind of task. This second task is Semantic Segmentation according to the Visual Object Classes (VOC)~\cite{PASCALVisualObject} protocol. For this task, the VOC~\cite{PASCALVisualObject} dataset is used. The splits used are the 2012 training and validation sets, and the 2007 test set. For this task we consider stitching one pair of networks, deeplabv3\_mobilenet\_v3\_large~\cite{chenRethinkingAtrousConvolution2017} and deeplabv3\_resnet50~\cite{chenRethinkingAtrousConvolution2017}.
The original weights were trained using both the ImageNet~\cite{russakovskyImageNetLargeScale2015} and Common Objects in Context (COCO)~\cite{linMicrosoftCOCOCommon2014,COCOCommonObjects} datasets.
Adam is used to train the stitching layers with a learning rate of $lr = 10^{-2}$ for $10$ epochs ($1464$ samples / epoch) - noting that training has generally converged at this point. Evaluation of a network is restricted to the first $64$ (out of $1449$) samples of the validation set to lower evaluation costs to the similar degree as ImageNet. Creation of this network took $791.37s$, using $46.17s$ for steps (1-3), and $745.20s$ to train the stitching layers (step 4). The resulting network has a total of $56$ matches. This provides a total of $\ell = 113$ decision variables. With the available computational resources for this experiment, we were able to evaluate up to $40$ solutions in parallel at a given time, with a batch size of $4$, evaluating up to $10$ solutions simultaneously per GPU.

\section{Experimental Setup}\label{sec:experimental-setup}
For ImageNet (a) and VOC, we apply all approaches in Section~\ref{sec:investigating-search-space}. For ImageNet (b), we only used GA and LK-GOMEA, the best two approaches according to the median on ImageNet (a) to save computational resources and time.
All runs are performed with an evaluation budget of 200,000 and a time limit of 24 hours, where the strictest limit applies. These limits were chosen as to allow for as thorough as search as possible.
We determine the population size for each approach by performing a single run for each $n \in [128, 256, 512, 1024, 2048]$ the chosen population size is then that of the configuration with the highest normalized hypervolume. Using this population size, we perform 5 runs per approach. 
Evaluations that have been skipped due to no change happening to active variables, as explained in Section~\ref{sec:stitching-nas}, still count towards the evaluation budget. The percentage of evaluations skipped will be discussed accordingly.
Experiments are either performed on (1) a cluster with 5 nodes. Where each node has 2x Intel Xeon Bronze 3206R CPU @ 1.90GHz, for a total of 16 CPU cores, 93~GB of RAM, and 3x NVIDIA RTX A5000 per machine. Or (2), a single node containing 2x Intel Xeon Platinum 8360Y (2x) @ 2.4 GHz, and 4x NVIDIA A100. All stitched networks were prepared on (1). The search for the ImageNet tasks were both performed on (1), while the segmentation task was performed on (2).
Evaluations performed during the optimization procedure are done using the validation set. The networks on the approximation front, as determined using accuracy on the validation set, will also have their accuracy determined on a test set. Summary results for hypervolume, using accuracy on the test set, will be included directly, detailed results using accuracy on both the validation and test set will be made available in the supplementary material. The best approach for each task, as determined by the median hypervolume, is compared against the other approaches using the Mann-Whitney U-test~\cite{mannTestWhetherOne1947,fayWilcoxonMannWhitneyTtestAssumptions2010} with $p=0.05$ and Holm-Bonferroni correction~\cite{holmSimpleSequentiallyRejective1979}.

The population size $n$ determined for each task / approach are as follows, where applicable. For ImageNet (a), $n_\textrm{GA} = 256$, $n_\textrm{GOMEA} = 512$ and $n_\textrm{LK-GOMEA} = 512$. For ImageNet (b), $n_\textrm{GA} = 128$ and $n_\textrm{LK-GOMEA} = 512$. For VOC,  $n_\textrm{GA} = 256$, $n_\textrm{GOMEA} = 2048$, and $n_\textrm{LK-GOMEA} = 2048$.

\section{Results and Discussion}\label{sec:results}

\begin{table*}[htb]
\centering
\caption{Summary statistics for hypervolume obtained by each approach on the test set of the corresponding task. Before processing, all fronts have networks with an accuracy lower a threshold $\text{accuracy}_\text{min}$ removed to avoid excessive contribution of networks performing significantly worse than the worst of the reference networks. This threshold is the accuracy of the worst performing network rounded down to the nearest multiple of 10. For ImageNet (a and b), $\text{accuracy}_\text{min} = 0.7$. For VOC, $\text{accuracy}_\text{min} = 0.9$. Using the reference point determined by the worst objective over the fronts $- 0.005\%$, and normalizing each axis by its range over the fronts $+ 0.01\%$. The approach with the highest median or not statistically different are marked in bold. For each combination, the 10th ($Q_{10}$) and 90th $Q_{90}$ quantiles and the median hypervolume are listed. }
\begin{tabular}{@{}rlllclllll@{}}
    \toprule
    \multicolumn{1}{c}{\textbf{task}} & \multicolumn{3}{c}{\textbf{ImageNet   (a)}} & \multicolumn{3}{c}{\textbf{ImageNet   (b)}} & \multicolumn{3}{c}{\textbf{VOC}} \\ \midrule
    \multicolumn{1}{r|}{\textbf{approach}} & \multicolumn{1}{c}{\textbf{$\mathbf{Q_{10}}$}} & \multicolumn{1}{c}{\textbf{median}} & \multicolumn{1}{c|}{\textbf{$\mathbf{Q_{90}}$}} & \textbf{$\mathbf{Q_{10}}$} & \multicolumn{1}{c}{\textbf{median}} & \multicolumn{1}{c|}{\textbf{$\mathbf{Q_{90}}$}} & \multicolumn{1}{c}{\textbf{$\mathbf{Q_{10}}$}} & \multicolumn{1}{c}{\textbf{median}} & \multicolumn{1}{c}{\textbf{$\mathbf{Q_{90}}$}} \\ \midrule
    \multicolumn{1}{r|}{\textbf{GA}} & \textbf{0.3494} & \textbf{0.3535} & \multicolumn{1}{l|}{\textbf{0.3563}} & \multicolumn{1}{l}{\textbf{0.8211}} & \textbf{0.8587} & \multicolumn{1}{l|}{\textbf{0.8711}} & \textbf{0.5779} & \textbf{0.5788} & \textbf{0.5792} \\
    \multicolumn{1}{r|}{\textbf{GOMEA}} & \textbf{0.3431} & \textbf{0.3452} & \multicolumn{1}{l|}{\textbf{0.3609}} & \multicolumn{1}{r}{-} & \multicolumn{1}{r}{-} & \multicolumn{1}{r|}{-} & \textbf{0.5773} & \textbf{0.5784} & \textbf{0.5787} \\
    \multicolumn{1}{r|}{\textbf{LK-GOMEA}} & \textbf{0.3519} & \textbf{0.3593} & \multicolumn{1}{l|}{\textbf{0.3631}} & \multicolumn{1}{l}{\textbf{0.8411}} & \textbf{0.8551} & \multicolumn{1}{l|}{\textbf{0.8698}} & \textbf{0.5781} & \textbf{0.5783} & \textbf{0.5788} \\
    \multicolumn{1}{r|}{\textbf{reference}} & \multicolumn{3}{c|}{0.2919} & \multicolumn{3}{c|}{0.7978} & \multicolumn{3}{c}{0.5488} \\ \bottomrule
\end{tabular}

\end{table*}

\begin{figure*}[tbph]
  \includegraphics[width=\textwidth]{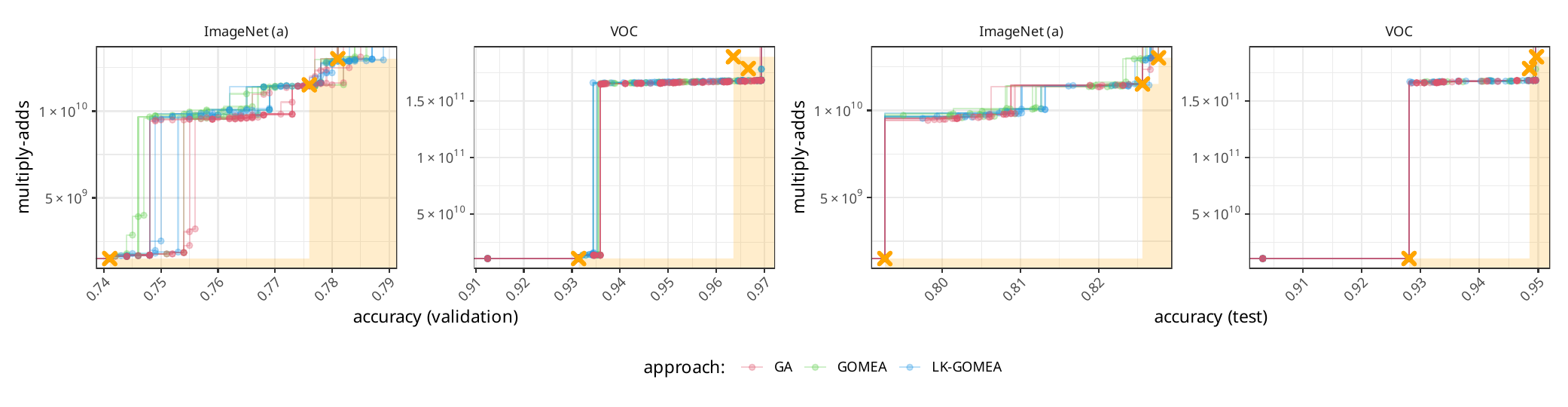}
  \caption{Approximation fronts obtained in each individual run, evaluated on the validation set as evaluated during a run (left) and evaluated on the full test set (right). The parents networks and their ensemble (referred to as 'reference networks') are labeled with an 'x', with the region in which networks dominate them marked in light orange. }
  \label{fig:fronts-tasks}
\end{figure*}

\begin{figure}[tbph]
  \includegraphics[width=\columnwidth]{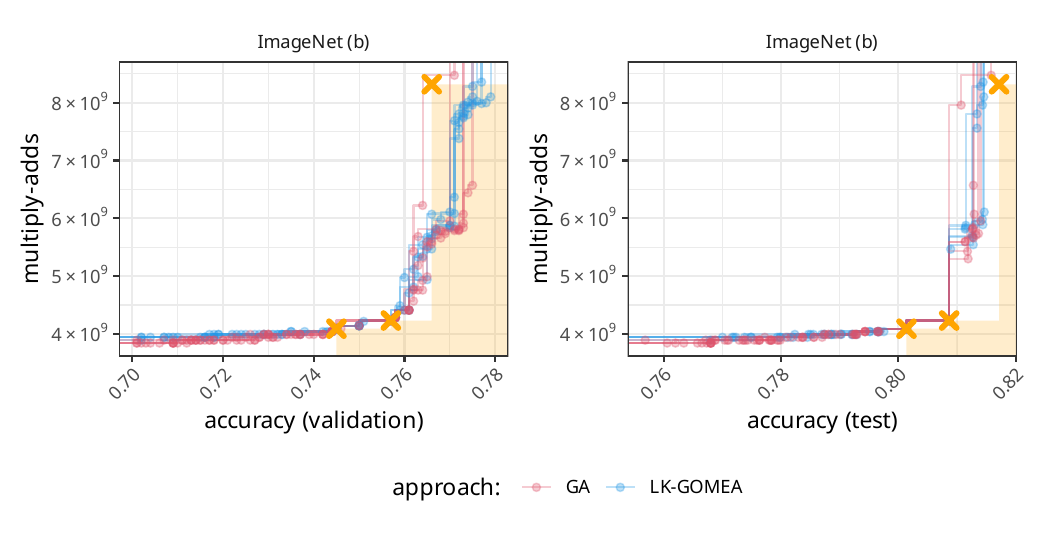}
  \caption{Fronts for ImageNet (b). Similar to Figure~\ref{fig:fronts-tasks}.}
  \label{fig:fronts-task-imagenet-b}
\end{figure}

\paragraph{Comparing approaches}
First, the only approaches in which the evaluations were skipped due to all changes being restricted to inactive variables, were GOMEA (with a median $45\%$ of evaluations being skipped) and LK-GOMEA (with $39\%$).
This has resulted in the runs of GOMEA and LK-GOMEA completing much earlier than those of the GA. 
This high percentage is due to the structure of the linkage tree, which contains many small subsets. In fact, half of the subsets form the univariate decomposition of the problem. As such, the difference with the reference solution is often small, and therefore has a much higher likelihood to consist of only inactive variables. Therefore, while not universal, skipping evaluation if only inactive variables have changed can be a significant improvement for approaches which make small changes.

For both ImageNet (a) and (b), all approaches excluding for LK-GOMEA were limited by their evaluation budget. LK-GOMEA terminates due to converged neighborhoods after using a median of $83\%$ of the budget for ImageNet (a) and $93\%$ for ImageNet (b). For VOC time was a limiting factor for the GA, with $94\%$ of the evaluation budget being spent in this case. GOMEA and LK-GOMEA were bounded by the number of evaluations, likely due to the skipping of evaluations saving time.

Reviewing the fronts obtained in Figure~\ref{fig:fronts-tasks} and \ref{fig:fronts-task-imagenet-b}, it is clear that the GA is capable of finding networks on the approximation front no other approach is able to find. Yet, the higher performance of these networks does not necessarily translate to an improvement on the test set, making performance differences between the approaches less important.
However, the desired speedup from linkage learning in GOMEA and LK-GOMEA seems to be absent. Based on the subsets of changed variables where evaluations were skipped, we suspect that the elements in the linkage tree often group inactive variables together.
This would not be surprising, as variables can take on any value when inactive, but are subject to selection pressure when active.
This increases the mutual information between pairs of variables where one activates the other, i.e., variables on the same path.
For an approach like GOMEA, this linkage is highly relevant. Any change to a variable, that activates another variable, especially one which has been subject to less selection pressure, can make the network worse. A simple local searcher altering a single variable at a time would likely get stuck. By sampling both variables jointly, an improvement could still be found. We suspect that the complex structure of variables activating and deactivating each other is hard to capture in a linkage tree, or even multiple linkage trees learned over a neighborhood, while two-point crossover implicitly respects such relationships due to the sequential nature of neural networks.

\paragraph{Found Networks} In Figures~\ref{fig:fronts-tasks} and \ref{fig:fronts-task-imagenet-b}, we have plotted the approximation fronts and the boundary of the dominated space obtained in each individual run.
For all tasks, we find that the cheapest reference network is not dominated. Networks with a lower computational cost and higher accuracy than this parent network are difficult to obtain in this search space. For example, picking the computationally cheapest layers between the two networks may require many stitches, yet simultaneously, the use of these stitches also tends to reduce performance. Future work may want to investigate further improvements to training of the stitches to decrease this performance reduction. Alternatively, a network may need to be constructed using a different matching, i.e. that the smallest network is also allowed to shrink. On the validation set, for ImageNet (a) and VOC, EAs are capable of finding networks that dominate the other parent network and the ensemble of both parent networks.

Yet, Finding an improvement that dominates one of the two parent networks can be difficult. This is especially the case if the two parent networks are similar in computational cost, as can be seen for ImageNet (b). This provides only little margin with respect to how much computational cost can be saved, while keeping performance high. Creating an ensemble with a shared portion of the two networks turns out to be an effective strategy to improve performance, without the full cost of an ensemble.

While networks may dominate with respect to the accuracy on the validation set, this accuracy is also an objective which has been for optimized by the approaches, making overfitting a concern.
When re-evaluating the networks on the fronts on test set, and filtering out the dominated networks yet again, the improvements in accuracy seem to disappear for the cheaper networks, while the more expensive networks do generalize to the test set and remain on the approximation front. While we note that one of the parent networks for ImageNet (a), and a parent network and the ensemble for VOC, remain dominated; the gap of these solutions to the reference networks, the parents and their ensemble, has shrunk considerably.
Improving the accuracy further, to increase this gap again, may be as simple as training the stitches of the networks of interest again. A brief investigation of this is included in the supplementary material.


If we compare the fronts obtained on ImageNet (a) and VOC in Figure~\ref{fig:fronts-tasks} against those obtained for ImageNet (b) in Figure~\ref{fig:fronts-task-imagenet-b}, it can be observed that the distribution of found networks is quite different. For the first two the front is noticeably concave, whereas on ImageNet (b) the front is mostly convex. How two networks are matched in steps 1 and 2, has a significant effect on the distribution of multiply-adds of the subnetworks that can be found. When two networks with greatly differing depth are stitched, as is the case for ImageNet (a) and VOC, a large portion of layers in the deeper network may remain without a match in the other network. If these layers are consecutive, all of these layers will either be included or excluded as one unit. This can result in large portion of the computational costs being present or absent, with no potential networks in between.
The approach used to match therefore has a large impact on the distribution of multiply-adds. The approach used in this work, matched as early as possible given the possible matches. In effect, for each group of compatible layers a large block is created. This resulted in the staircase for ImageNet (a), and the large cliff for VOC. Altering the matching approach for a better distribution of matches may therefore be necessary to obtain a less concave front.

Overfitting to accuracy is possible with minor increases in predictions, i.e. such that the predicted class becomes the true class. This would appear in the calibration of the predictions of the network, i.e., how well does the predicted probability represent the actual probability that the classification is correct. Furthermore, good calibration may be desirable for many real-world applications. We have therefore additionally investigated the impact on the argmax calibration~\cite{guoCalibrationModernNeural2017,mindererRevisitingCalibrationModern2021a} of the networks using the Expected Calibration Error (ECE)~\cite{naeiniObtainingWellCalibrated2015} for the best networks found on ImageNet (a). To compute ECE the probabilities of the predicted class are binned, and for each bin the accuracy is calculated. If a network is calibrated, one would expect samples in the $0.3-0.4$ bin to be predicted correctly roughly $35\%$ of the time. For some of the networks, we observe for the lower bins that the actual accuracy is much higher than what one would expect. These networks are therefore underconfident. This is in line with what would be expected if accuracy is maximized. Yet, as this does not hold for all networks, differences in the data distribution between ImageNet and ImageNetV2 may also be a cause. Future work may want to consider using cross-entropy loss over accuracy, especially when calibration is important.

Finally, a significant subset of the networks found by our approach are an ensemble with a portion of the computational work shared between the two networks. This often provides similar benefits as those obtained by normal ensembles, for example, the resulting networks tend to have a lower calibration error (ECE) than the original parent networks. Yet, their computational cost is still reduced compared to the full ensemble.

Given the aforementioned results we have shown that it is possible to recombine neural networks, without task-specific training, into a supernetwork, and obtain offspring that are functional. While the resulting supernetwork may be structurally complex, the resulting offspring can provide a novel trade-off when compared the parent networks and their ensemble.

\section{Conclusion}\label{sec:conclusion}
In this work, we have investigated model stitching as an approach for recombining two neural networks while preserving their original performance. The technique can employ already trained neural networks, and construct many new offspring networks without requiring additional training for each offspring. This paves the way for neuroevolution techniques that can employ both pre-training and crossover, allowing them to be applied to problems with fewer data points, while also providing an alternative pathway for distributed training of neural networks. Using parallel EAs to search through the space of all possible recombinations, we found that recombined networks can provide a novel trade-off between the performance and computational cost of the original parent networks, and potentially even dominate parent networks.

\begin{acks}
This publication is part of the project "DAEDALUS - Distributed and Automated Evolutionary Deep Architecture Learning with Unprecedented Scalability" with project number \grantnum{NWO}{18373} of the research programme \grantnum{NWO}{Open Technology Programme} which is (partly) financed by the \grantsponsor{NWO}{Dutch Research Council (NWO)}{}. Other financial contributions as part of this project have been provided by 
\grantsponsor{CElekta}{Elekta AB}{}
 and 
\grantsponsor{CORTECLC}{Ortec Logiqcare B.V.}{}. Furthermore, we thank NWO for the Small Compute grant on the Dutch National Supercomputer Snellius.
\end{acks}

\bibliographystyle{ACM-Reference-Format}
\bibliography{2023-11-StitchingForNeuroevolution.bib}

\end{document}